\documentclass[11pt]{article}
\usepackage[utf8]{inputenc}
\usepackage[T1]{fontenc}
\usepackage{amsmath,amssymb}
\usepackage{graphicx}
\usepackage{booktabs}
\usepackage{url}
\usepackage{cite}
\usepackage{geometry}
\usepackage{microtype}

\def\BibTeX{{\rm B\kern-.05em{\sc i\kern-.025em b}\kern-.08em
T\kern-.1667em\lower.7ex\hbox{E}\kern-.125emX}}

\begin{document}
\title{When More References Hurt: Contamination-Aware DINOv2 Memory Banks for Few-Shot Steel Defect Detection}
\author{
Hannaneh Kalantary\\
\textit{University of Padua}\\
\texttt{hannaneh.kalantary@studenti.unipd.it}
\and
Javad Khoramdel\\
\textit{University of Vienna}\\
\texttt{j.khorramdel96@gmail.com}
}
\maketitle

\begin{abstract}
Patch-memory anomaly detectors assume that their reference bank is normal, an assumption that is difficult to guarantee when additional industrial images are unverified. We study whether a few trusted normal images can safely recover useful normal patches from such references without defect masks. Starting from the DINOv2 patch-memory formulation used by AnomalyDINO, we score candidate patches by distance to a clean seed bank, discard the most suspicious 20\%, merge the retained patches with the seed, and enforce a fixed budget by greedy coreset selection. On Severstal, naive additional references contain 9.46\% anomalous patches; the proposed trim rejects 78.1\% of them and reduces residual contamination to 2.59\%. At an equal 51,200-patch development budget, the proposed bank reaches 0.1084 AUPRC versus 0.0950 for naive expansion, 0.0952 for random removal, and 0.1030 for eight clean images. Injecting only 0.5\% anomalous patches into a clean bank reduces AUPRC from 0.1030 to 0.0759. On all five completed held-out pairs, the proposed bank improves over naive expansion, with a mean gain of 0.0142 AUPRC. Reference purity is therefore a first-order design variable, and unverified images are useful only when their contribution is filtered explicitly.
\end{abstract}

\noindent\textbf{Keywords---} Anomaly Detection, Few-shot Learning, Few-shot Detection, Steel surface defects, DINOv2, Foundational Model

\section{Introduction}
Automated steel-surface inspection is commonly formulated as supervised classification, detection, or segmentation. Komijani et al. study multi-label recognition with transfer learning and Vision Transformers~\cite{komijani2022steel}; Sabet et al. evaluate U-Net and FCN variants for pixel-level segmentation~\cite{sabet2022steel}; and Ashrafi et al. combine detection and segmentation to improve localization of small defects~\cite{ashrafi2025steel}. These approaches show that modern networks can model steel defects effectively when labeled examples and spatial annotations are available, but obtaining such supervision is costly when defects are rare, subtle, or continually changing.

The broader shift toward transformer representations also motivates lower-supervision inspection. Lightweight ViT and hybrid CNN--ViT models transfer effectively to plant-disease recognition~\cite{borhani2022plant}, while DINOv2 provides generic self-supervised visual features that transfer without task-specific fine-tuning~\cite{oquab2024dinov2}. Industrial anomaly detection exploits this transferability by learning normality rather than enumerating every possible defect. PaDiM models distributions of pretrained patch features~\cite{defard2021padim}, PatchCore stores a representative memory of normal patch embeddings~\cite{roth2022patchcore}, and recent foundation-model approaches exploit CLIP or DINOv2 representations~\cite{jeong2023winclip,zhou2024anomalyclip,damm2025anomalydino}. AnomalyDINO is especially relevant here because frozen DINOv2 features and nearest-neighbor patch memory make reference composition a direct part of the detector.

That simplicity creates a deployment risk: every stored patch becomes part of the definition of normality. Benchmarks such as MVTec AD and VisA provide curated nominal training sets~\cite{bergmann2019mvtec,zou2022visa}, whereas production lines may provide many images that are cheap to collect but expensive to verify. A small defect region inside an otherwise useful reference can therefore insert misleading embeddings into the memory. Discarding every uncertified image wastes normal appearance variation, but ingesting it naively can teach the detector that defects are normal.

We therefore ask: \emph{can a few trusted normal images safely mine normal patches from additional unverified images?} Using Severstal~\cite{severstal2019} and a fixed DINOv2 patch-memory detector, we (i) quantify sensitivity to controlled reference contamination, (ii) introduce a mask-free expansion strategy that rejects the 20\% of candidate patches most distant from a trusted seed, (iii) separate purification quality from memory size with oracle, random, clean-shot, and exact-budget controls, and (iv) report development and currently available held-out results while explicitly excluding incomplete experimental branches from the evidence.

\section{Related Work}
\subsection{Supervised Steel Inspection}
Steel defect inspection spans classification, detection, and segmentation. Komijani et al. use supervised transfer learning and ViT models for multi-label steel defect recognition~\cite{komijani2022steel}. Sabet et al. evaluate U-Net variants and FCN-8 for pixel-level defect segmentation~\cite{sabet2022steel}, while Ashrafi et al. compare detection and segmentation and combine YOLOv4 localization with U-Net refinement~\cite{ashrafi2025steel}. These methods demonstrate strong defect modeling when labels are available, but classification requires categories, detection requires boxes, and segmentation requires masks. Our setting is complementary: only a small trusted normal set is assumed, and the objective is abnormal-patch localization without training a defect classifier.

\subsection{Foundation Features and Patch Memories}
Vision Transformers introduced patch-token processing as a general alternative to convolutional encoders~\cite{dosovitskiy2021vit}. Their transferability has been demonstrated across domains, including plant-disease recognition with lightweight ViT and hybrid CNN--ViT models~\cite{borhani2022plant}. DINOv2 further showed that large-scale self-supervised pretraining can provide generic visual features without task-specific fine-tuning~\cite{oquab2024dinov2}, which is attractive when industrial deployments provide few target-domain examples.

Patch-based anomaly detectors compare local test features with representations of nominal data. PaDiM fits location-dependent feature distributions~\cite{defard2021padim}, whereas PatchCore stores a coreset-reduced bank of normal embeddings~\cite{roth2022patchcore}. Few-shot and zero-shot methods reduce target supervision further: WinCLIP and AnomalyCLIP exploit vision-language priors~\cite{jeong2023winclip,zhou2024anomalyclip}, while AnomalyDINO uses frozen DINOv2 features and a simple nearest-neighbor patch memory~\cite{damm2025anomalydino}. Most such formulations optimize representation or compression assuming the nominal references are clean. We focus on a distinct failure mode: contamination already present in the reference images. Coreset selection can reduce redundancy, but it does not guarantee that anomalous patches are removed, motivating purification before the final memory budget is enforced.

\section{Method}
\label{sec:method}
\begin{figure*}[!t]
\centering
\includegraphics[width=0.9\textwidth]{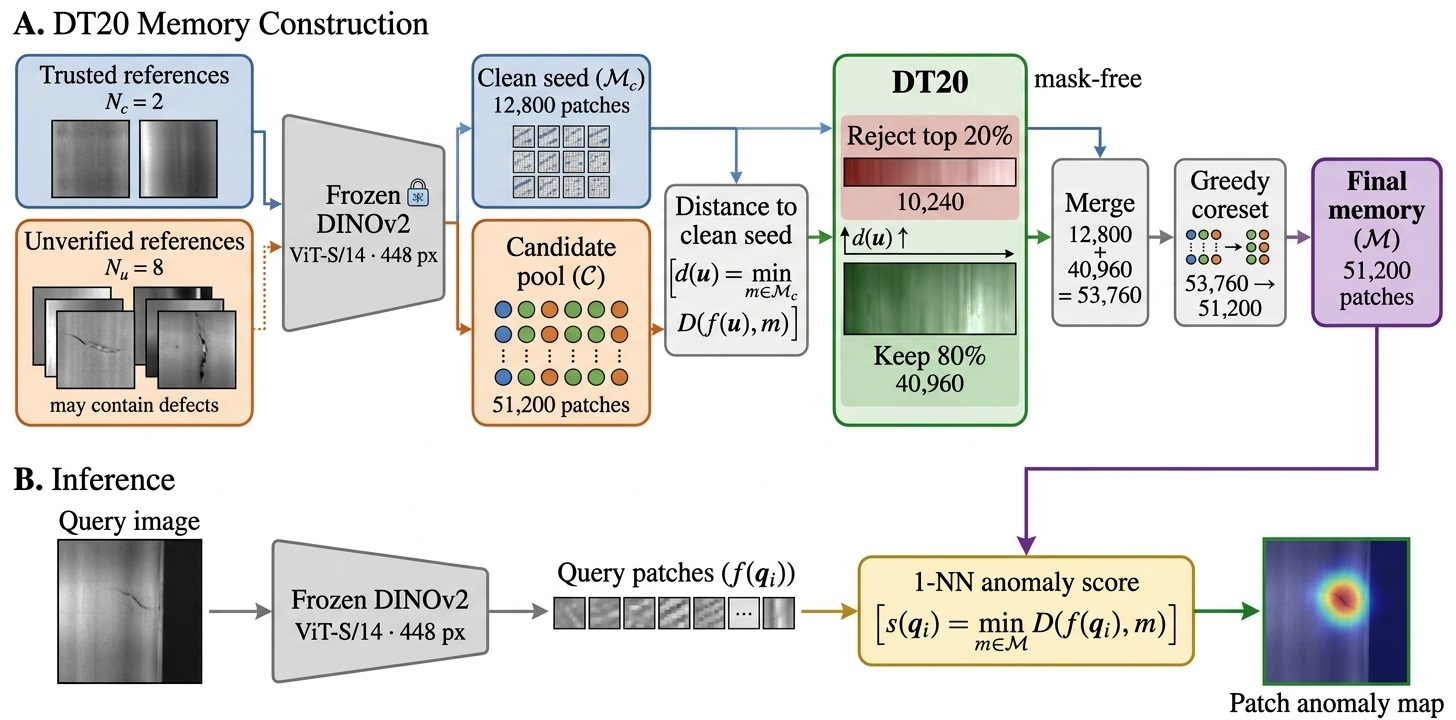}
\caption{DT20 construction and inference. Two trusted references form the clean seed memory; eight unverified references provide candidates. Each candidate is scored by nearest distance to the seed, the largest 20\% are rejected, and the retained patches are merged with the seed and reduced to a 51,200-patch memory by greedy coreset selection. Query patches are scored by nearest-memory distance. No defect masks are used by DT20.}
\label{fig:method-overview}
\end{figure*}

\subsection{Base Detector and Reference Expansion}
Let $f(x)\in\mathbb{R}^{H_p\times W_p\times d}$ denote the frozen DINOv2 patch features of image $x$. Following AnomalyDINO, the normal memory bank $\mathcal{M}$ stores reference patch embeddings. A query patch $q$ receives the nearest-neighbor anomaly score
\begin{equation}
s(q)=\min_{m\in\mathcal{M}}\lVert f(q)-m\rVert_2,
\end{equation}
so larger distance indicates poorer agreement with the normal appearance represented by the bank. We keep the backbone and scoring rule fixed and alter only how $\mathcal{M}$ is constructed.

The primary expansion setting contains a trusted clean set $\mathcal{R}_c$ and a larger unverified set $\mathcal{R}_u$. Two clean images contribute 12,800 seed embeddings and eight additional images contribute 51,200 candidates under our preprocessing. The additional images may contain both normal steel and local defects. A naive expansion inserts every candidate and therefore risks teaching the detector that reference defects are normal. For analysis only, an oracle removes every candidate that overlaps a ground-truth defect.

\subsection{Distance-Trimmed Purification}
DT20 uses the clean seed memory $\mathcal{M}_c$ to assess each candidate $u\in\mathcal{R}_u$:
\begin{equation}
d(u)=\min_{m\in\mathcal{M}_c}\lVert f(u)-m\rVert_2.
\end{equation}
Candidates are ranked by $d(u)$, and the largest fixed fraction $\rho$ is rejected. The retained set is
\begin{equation}
\mathcal{C}_{\rho}=\{u\in\mathcal{R}_u:d(u)\le Q_{1-\rho}(d)\},
\end{equation}
where $Q_{1-\rho}$ is the empirical distance quantile over the candidate pool. Development-fold controls select $\rho=0.20$, so the method keeps the 80\% of additional patches most compatible with the trusted seed. These retained candidates are merged with $\mathcal{M}_c$ and greedily coreset-reduced to the fixed budget $B=51{,}200$ whenever enough candidates are available. The frozen proposed configuration is therefore two clean seed images, eight class-balanced additional images, 20\% distance trimming, and an exact 51,200-patch final memory. We refer to it as \textbf{DT20}.

\subsection{Controls}
We use four controls to separate purification quality from memory size and reference count. The \emph{clean} control uses only verified defect-free images. The \emph{naive} control adds all candidate patches. \emph{Random20} removes 20\% of candidates uniformly at random, matching DT20's retention rate without using the distance score. The \emph{oracle} removes candidates with any ground-truth defect overlap and is reported only as an analysis upper bound. Exact-budget rows further distinguish targeted purification from improvements that might arise merely from memory capacity or greedy coreset selection.

\section{Experimental Setup}
\subsection{Dataset and Preprocessing}
We use the Severstal Steel Defect Detection dataset~\cite{severstal2019}, which contains $256\times1600$ steel-strip images and run-length-encoded masks for four defect classes. Images without defect annotations are treated as defect-free. The detector uses DINOv2 ViT-S/14 at a target resolution of 448 pixels on the shorter edge. Dimensions are cropped to multiples of the 14-pixel patch size, yielding 6,400 patch tokens per reference image in the primary configuration.

Ground-truth masks are mapped to the patch grid for evaluation. A patch is labeled anomalous when at least 50\% of its area overlaps a defect mask. These masks are never used by DT20; they are used only to compute evaluation metrics and to define oracle analyses.

\subsection{Development and Held-Out Protocol}
The repository defines five stratified folds. Fold 0 is used for method development and controlled studies, with reference-selection seeds 42 through 46 for replication. The 20\% trim fraction and 51,200-patch memory budget are frozen from fold-0 controls before held-out evaluation.

Available held-out results cover folds 1 and 2 with seeds 42, 43, and 44. Clean, naive, and random20 have six completed fold-seed cells. DT20 and oracle have five because their fold-2 seed-44 jobs are unavailable. We report means over completed cells and paired DT20-versus-naive differences only when both results exist; these partial experiments are not presented as a completed five-fold cross-validation. Exploratory SAM2 refinement, efficiency, and later anomaly-memory branches are also excluded from the claims, as are invalid class-specific contamination runs that inserted zero anomalous patches.

\subsection{Metrics}
Patch-level AUPRC is the primary metric because anomalous patches are strongly imbalanced. AUROC and maximum F1 over thresholds (F1-max) provide supporting ranking and operating-point views. Fixed-threshold F1 is retained in selected tables for completeness but varies strongly with calibration, so method selection and the main claims rely on AUPRC, supported by AUROC and F1-max.

\section{Results}
\subsection{Purification Quality and Trim Selection}
In the two-clean plus eight-additional development setting, 4,845 of 51,200 candidate patches are anomalous under an any-overlap oracle criterion, or 9.46\%. Table~\ref{tab:purity} shows that DT20 rejects 78.1\% of anomalous candidates while retaining 86.1\% of normal candidates, reducing residual contamination to 2.59\%. Random removal leaves contamination essentially unchanged.

\begin{table}[!t]
\caption{Purification quality on fold 0, seed 42.}
\label{tab:purity}
\centering
\resizebox{\columnwidth}{!}{%
\begin{tabular}{lrrrr}
\toprule
Bank & Reject recall & Final contam. & Normal retain & Reject prec. \\
\midrule
Naive & 0.000 & 0.0946 & 1.000 & 0.000 \\
Random size-matched & 0.195 & 0.0953 & 0.799 & 0.092 \\
DT20 & \textbf{0.781} & \textbf{0.0259} & 0.861 & 0.369 \\
Oracle & 1.000 & 0.0000 & 1.000 & 1.000 \\
\bottomrule
\end{tabular}}
\end{table}

The distance rule removes 78.1\% of anomalous candidates while retaining 86.1\% of normal ones. Its reject precision is only 0.369, so some unusual but normal patches are deliberately sacrificed for safety. Random size-matched removal rejects a similar amount of data yet leaves contamination at 9.53\%, showing that the useful signal comes from compatibility with the trusted seed rather than from simply shrinking the candidate pool.

Because DT20 intentionally sacrifices some normal diversity to reduce contamination, the trim rate is selected on development data. Fixed trims of 5\%, 10\%, and 20\% reach AUPRC values of 0.0987, 0.1022, and 0.1084, respectively. The strongest percentile-based automatic rule, a 95th-percentile acceptance threshold, reaches 0.1007, while random size-matched removal remains near 0.0951. The 20\% setting also gives 0.8132 AUROC and 0.2151 F1-max, so we freeze it before held-out evaluation rather than retuning per fold.

\begin{table}[!t]
\caption{Fold-0 purification controls used to freeze the trim rate.}
\label{tab:trim}
\centering
\begin{tabular}{lccc}
\toprule
Setting & AUPRC & AUROC & F1-max \\
\midrule
Random size-matched & 0.0951 & 0.7971 & 0.1935 \\
Fixed trim 5\% & 0.0987 & 0.8020 & 0.1986 \\
Fixed trim 10\% & 0.1022 & 0.8062 & 0.2039 \\
Auto percentile 95 & 0.1007 & 0.8045 & 0.2015 \\
Fixed trim 20\% & \textbf{0.1084} & \textbf{0.8132} & \textbf{0.2151} \\
\bottomrule
\end{tabular}
\end{table}

\begin{figure*}[!t]
\centering
\begin{tabular}{@{}c@{}}
\textbf{(a) Unverified reference image} \\[2pt]
\includegraphics[width=0.8\textwidth]{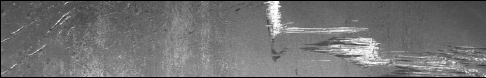} \\[5pt]
\textbf{(b) Ground-truth defect mask} \\[2pt]
\includegraphics[width=0.8\textwidth]{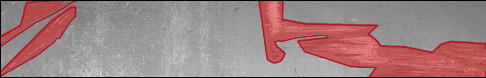} \\[5pt]
\textbf{(c) Patch distance to clean seed} \\[2pt]
\includegraphics[width=0.8\textwidth]{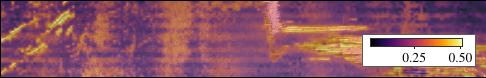} \\[5pt]
\textbf{(d) Rejected 20\%} \\[2pt]
\includegraphics[width=0.8\textwidth]{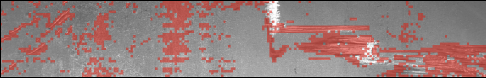} \\[5pt]
\textbf{(e) Retained 80\%} \\[2pt]
\includegraphics[width=0.8\textwidth]{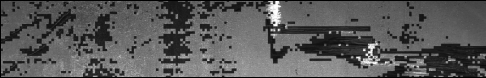}
\end{tabular}
\caption{Qualitative DT20 purification on a representative unverified reference image. (a) Input image; (b) ground-truth defect mask, shown only for retrospective analysis and never seen by DT20; (c) per-patch distance $d_u$ to the clean seed bank; (d) the 20\% of patches rejected by the distance trim; (e) the retained 80\% merged into the memory. Rejected high-distance patches in (d) concentrate around the true defect region in (b), consistent with the 78.1\% reject recall reported in Table~\ref{tab:purity}.}
\label{fig:qualitative}
\end{figure*}

\subsection{Exact Memory-Budget Controls}
Table~\ref{tab:budget} compares key conditions at the same 51,200-patch budget. This control matters because a larger reference set can improve nearest-neighbor coverage simply by creating a larger memory. Naive two-plus-eight expansion, randomly reduced to the target budget, reaches 0.0950 AUPRC, below clean-8 at 0.1030. Random20 with greedy reduction remains at 0.0952, whereas DT20 reaches 0.1084, gains of 0.0134 over naive and 0.0054 over clean-8 on fold 0. Oracle reaches 0.1072, close to DT20 in this run. With four clean seed images plus eight additional images, DT20 reaches the strongest completed fold-0 result, 0.1140.

These results support two separate conclusions. First, additional images are not automatically helpful: unfiltered expansion is worse than a clean bank at the same final capacity. Second, DT20's improvement is not a memory-size artifact. Naive random-budget matching (0.0950) and random20 with greedy matching (0.0952) are nearly identical despite using different budget mechanisms. Greedy coreset selection therefore contributes little when the underlying patch selection is not targeted at contamination; the useful change is which candidate patches survive before compression.

\begin{table*}[!t]
\caption{Fold-0 reference composition and matched-budget controls.}
\label{tab:budget}
\centering
\begin{tabular}{lccrcccc}
\toprule
Method & Clean & Add. & Final bank & AUPRC & AUROC & F1-max & Fixed F1 \\
\midrule
Clean 1 & 1 & 0 & 6,400 & 0.0700 & 0.703 & 0.135 & 0.062 \\
Clean 2 & 2 & 0 & 12,800 & 0.1020 & 0.818 & 0.191 & 0.181 \\
Clean 4 & 4 & 0 & 25,600 & 0.1050 & 0.813 & 0.200 & 0.177 \\
Clean 8 & 8 & 0 & 51,200 & 0.1030 & 0.795 & 0.199 & 0.147 \\
Naive, random budget & 2 & 8 & 51,200 & 0.0950 & 0.797 & 0.193 & 0.080 \\
Random20 + greedy & 2 & 8 & 51,200 & 0.0952 & 0.7974 & 0.1936 & 0.0274 \\
Oracle + greedy & 2 & 8 & 51,200 & 0.1072 & 0.8109 & 0.2124 & 0.1200 \\
DT20 + greedy & 2 & 8 & 51,200 & \textbf{0.1084} & \textbf{0.8132} & \textbf{0.2151} & 0.0620 \\
DT20 + greedy & 4 & 8 & 51,200 & \textbf{0.1140} & 0.815 & 0.218 & 0.136 \\
\bottomrule
\end{tabular}
\end{table*}

\subsection{Multi-Seed Repeatability on the Development Fold}
A separate five-seed fold-0 study, conducted before DT20 was frozen, tests sensitivity to which references are selected. Table~\ref{tab:multiseed} reports means and standard deviations. Naive expansion is negative relative to clean-2 in four of five seeds, while the earlier percentile rule improves over naive in four seeds with one tie. Oracle purification is consistently stronger in ranking metrics. Because this study predates the final fixed 20\% rule, it establishes the general reference-composition trend rather than DT20's own replicated performance.

\begin{table}[!t]
\caption{Five-seed fold-0 reference-composition study.}
\label{tab:multiseed}
\centering
\resizebox{\columnwidth}{!}{%
\begin{tabular}{lcccc}
\toprule
Condition & AUPRC & AUROC & F1-max & Fixed F1 \\
\midrule
Clean 2 & $0.102\pm0.017$ & $0.784\pm0.030$ & $0.189\pm0.027$ & $0.098\pm0.099$ \\
Naive 2+8 & $0.101\pm0.011$ & $0.796\pm0.020$ & $0.198\pm0.017$ & $0.110\pm0.025$ \\
Percentile-95 purified & $0.103\pm0.012$ & $0.797\pm0.020$ & $0.200\pm0.019$ & $0.027\pm0.037$ \\
Oracle purified & $\mathbf{0.110}\pm0.011$ & $\mathbf{0.804}\pm0.020$ & $\mathbf{0.211}\pm0.017$ & $0.142\pm0.021$ \\
\bottomrule
\end{tabular}}
\end{table}

\subsection{Controlled Contamination}
To isolate contamination from capacity, we start from a clean eight-image bank of 51,200 patches and replace a controlled fraction with anomalous patches while keeping bank size fixed. Table~\ref{tab:contam} shows a sharp initial failure: replacing only 256 entries (0.5\%) reduces AUPRC from 0.1030 to 0.0759, a 26.3\% relative drop. Performance then decreases more gradually, reaching 0.0587 at 20\% contamination. Because capacity is constant throughout, this experiment directly attributes the loss to anomalous memory content rather than to bank size. Uniform and class-balanced sampling produced the same values in this run. Invalid single-class arms that inserted zero anomalous patches are excluded because they reproduce the clean-bank metric and do not support class-specific sensitivity claims.

\begin{figure}[!t]
\centering
\includegraphics[width=\columnwidth]{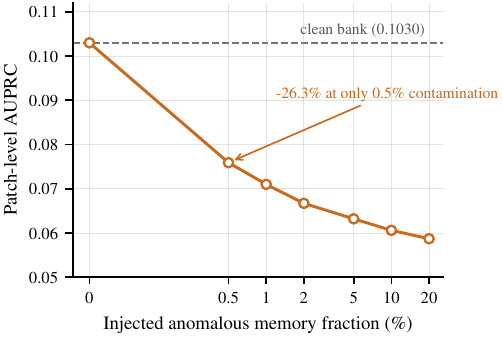}
\caption{AUPRC under controlled anomaly replacement at fixed 51,200-patch memory size.}
\label{fig:contamination-curve}
\end{figure}

\begin{table}[!t]
\caption{Controlled replacement contamination.}
\label{tab:contam}
\centering
\begin{tabular}{@{}ccc@{}}
\toprule
Contamination & Injected patches & AUPRC \\
\midrule
0 & 0 & 0.1030 \\
0.5\% & 256 & 0.0759 \\
1\% & 512 & 0.0710 \\
2\% & 1,024 & 0.0667 \\
5\% & 2,560 & 0.0632 \\
10\% & 5,120 & 0.0606 \\
20\% & 10,240 & 0.0587 \\
\bottomrule
\end{tabular}
\end{table}

\subsection{Held-Out Results}
Table~\ref{tab:heldout} summarizes the available held-out cells. Clean-8 has the highest completed-cell mean AUPRC (0.1406), an important counterpoint to the fold-0 result: collecting additional verified clean references remains the safest strategy when such references are available. DT20 is intended for the harder regime in which additional images can be collected but cannot be exhaustively certified.

For that regime, the most reliable held-out comparison is paired DT20 versus naive expansion. Across the five fold-seed pairs where both are available, the AUPRC gains are $+0.0068$, $+0.0276$, $+0.0129$, $+0.0055$, and $+0.0183$. All five are positive, with mean $+0.0142$. Random20 tracks naive expansion, again indicating that targeted filtering matters more than discarding patches indiscriminately. Because DT20 and oracle lack one fold-2 cell, their aggregate means should not be compared as though they were based on a complete cross-validation matrix.

\begin{table}[!t]
\caption{Available held-out results on folds 1--2. DT20/oracle use five completed cells; others use six.}
\label{tab:heldout}
\centering
\scriptsize
\resizebox{\columnwidth}{!}{%
\begin{tabular}{lccccc}
\toprule
Condition & $n$ & AUPRC mean [95\% CI] & AUROC & F1-max & Fixed F1 \\
\midrule
Clean 2 & 6 & 0.0979 [0.085, 0.111] & 0.780 & 0.187 & 0.063 \\
Clean 8 & 6 & \textbf{0.1406 [0.115, 0.173]} & \textbf{0.823} & \textbf{0.234} & 0.112 \\
Naive 2+8 & 6 & 0.0919 [0.085, 0.099] & 0.781 & 0.180 & 0.111 \\
Random20 2+8 & 6 & 0.0921 [0.085, 0.099] & 0.782 & 0.181 & 0.010 \\
DT20 2+8 & 5 & 0.1062 [0.100, 0.113] & 0.790 & 0.203 & 0.038 \\
Oracle 2+8 & 5 & 0.0998 [0.090, 0.109] & 0.791 & 0.193 & \textbf{0.140} \\
\bottomrule
\end{tabular}}
\end{table}

\section{Discussion}
The experiments identify reference purity as a first-order variable in patch-memory anomaly detection. A nearest-neighbor detector has no mechanism to distinguish a mislabeled anomalous memory patch from a true normal patch once both are stored. The controlled contamination study shows that the resulting failure is not gradual: even 0.5\% replacement causes a large AUPRC drop. This explains why naive expansion can fail although most pixels in a defective steel image are visually normal.

DT20 is intentionally simple. It does not segment a reference defect or predict a defect class; instead, the clean seed bank provides a local definition of normality, and candidate patches least compatible with that seed are withheld. The purification analysis also makes the trade-off explicit. A 20\% trim rejects 78.1\% of anomalous candidates but has reject precision 0.369, so some unusual normal patches are sacrificed. Nevertheless, residual contamination falls sharply, and matched-budget controls show that the gain is not explained by a larger bank or by coreset selection alone.

The method therefore trades reference certainty for appearance coverage. Two trusted clean images may not span normal texture variation; additional images contribute diversity after suspicious regions are removed. The 4-clean plus 8-additional result, which reaches 0.1140 AUPRC on fold 0 at the same final budget, supports this interpretation: a stronger trusted seed can improve the expanded bank without increasing final memory capacity.

The held-out results also prevent an overly broad conclusion. Clean-8 remains stronger than DT20 2+8 on the currently available folds, so the proposed method should not be presented as a replacement for acquiring verified normal references when those are easy to obtain. The more specific deployment rule is to prefer a clean-only bank when enough certified images exist, and to use contamination-aware expansion when the trusted seed is small but additional unverified imagery is plentiful. This conditional claim is directly aligned with industrial data collection, where image acquisition can be cheaper than exhaustive verification.

\section{Limitations}
The held-out matrix is incomplete: evidence covers only folds 1 and 2, and one DT20/oracle cell is missing for each method, so the results should not be interpreted as completed five-fold cross-validation. The 20\% trim rate is selected on a single development fold and may not be optimal under different steel lines, illumination, or acquisition conditions. We evaluate one DINOv2 backbone and one nearest-neighbor scoring rule, so quantitative contamination sensitivity may differ for other feature representations or anomaly scores. Severstal's pixel masks make oracle contamination analysis possible, whereas residual contamination cannot be measured directly in deployment. Fixed-threshold F1 also remains unstable across reference compositions, indicating that a production system would require explicit operating-point calibration. Finally, incomplete SAM2 refinement, efficiency, anomaly-memory, and attention branches are deliberately excluded rather than used to strengthen the claims.

\section{Conclusion}
More reference images can hurt patch-memory anomaly detection when their patches are not guaranteed to be normal. On Severstal, only 0.5\% controlled anomaly contamination sharply reduces AUPRC, and naive ingestion of unverified images underperforms a clean bank at matched size. DT20 uses a small trusted seed to filter candidate patches, substantially reduces contamination, and improves over naive expansion on every completed held-out pair. Verified clean references remain the safest option when available; when they are scarce, contamination-aware expansion can recover useful normal appearance from otherwise risky imagery.

\end{document}